\documentclass[conference]{IEEEtran}
\IEEEoverridecommandlockouts
\usepackage{cite}
\usepackage{amsmath,amssymb,amsfonts}
\usepackage{algorithmic}
\usepackage{graphicx}
\usepackage{textcomp}
\usepackage{xcolor}
\usepackage{multirow}
\usepackage{booktabs}
\usepackage{siunitx}
\usepackage{subfigure}
\usepackage[colorlinks=true,linkcolor=red,citecolor=blue]{hyperref}
\usepackage{tikz}

\def\BibTeX{{\rm B\kern-.05em{\sc i\kern-.025em b}\kern-.08em
    T\kern-.1667em\lower.7ex\hbox{E}\kern-.125emX}}

\newcommand\copyrighttext{%
  \footnotesize \textcopyright \the\year{} IEEE. Personal use of this material is permitted.  Permission from IEEE must be obtained for all other uses, in any current or future media, including reprinting/republishing this material for advertising or promotional purposes, creating new collective works, for resale or redistribution to servers or lists, or reuse of any copyrighted component of this work in other works.}

\newcommand\copyrightnotice{%
\begin{tikzpicture}[remember picture,overlay]
\node[anchor=south,yshift=10pt] at (current page.south) {\fbox{\parbox{\dimexpr0.75\textwidth-\fboxsep-\fboxrule\relax}{\copyrighttext}}};
\end{tikzpicture}%
}

\begin{document}

\title{A Closer Look at Data Augmentation Strategies for Finetuning-Based Low/Few-Shot Object Detection \\
\thanks{This project has received funding from the European Union’s Horizon Europe research and innovation programme under grant agreement No. 101070181 (TALON).}
}

\author{Vladislav Li\IEEEauthorrefmark{1}, Georgios Tsoumplekas\IEEEauthorrefmark{2}, Ilias Siniosoglou\IEEEauthorrefmark{2}\IEEEauthorrefmark{3}, Vasileios Argyriou\IEEEauthorrefmark{1}, Anastasios Lytos\IEEEauthorrefmark{4}, \\Eleftherios Fountoukidis\IEEEauthorrefmark{4} and Panagiotis Sarigiannidis\IEEEauthorrefmark{2}\IEEEauthorrefmark{3}

\thanks{\IEEEauthorrefmark{1} V. Li and V. Argyriou are with the Department of Networks and Digital Media, Kingston University, Kingston upon Thames, United Kingdom - \texttt{E-Mail: \{v.li, vasileios.argyriou\}@kingston.ac.uk}}

\thanks{\IEEEauthorrefmark{2} G. Tsoumplekas, I. Siniosoglou and P. Sarigiannidis are with the R\&D Department, MetaMind Innovations P.C., Kozani, Greece - \texttt{E-Mail: \{gtsoumplekas, isiniosoglou, psarigiannidis\}@metamind.gr}}

\thanks{\IEEEauthorrefmark{3} I. Siniosoglou and P. Sarigiannidis are with the Department of Electrical and Computer Engineering, University of Western Macedonia, Kozani, Greece - \texttt{E-Mail: \{isiniosoglou, psarigiannidis\}@uowm.gr}}

\thanks{\IEEEauthorrefmark{4} A. Lytos and E. Fountoukidis are with Sidroco Holdings Ltd., Nicosia, Cyprus - \texttt{E-Mail: \{alytos, efountoukidis\}@sidroco.com}}
}

% \author{Anonymous Authors}

\maketitle
\copyrightnotice

\begin{abstract}
Current methods for low- and few-shot object detection have primarily focused on enhancing model performance for detecting objects. One common approach to achieve this is by combining model finetuning with data augmentation strategies. However, little attention has been given to the energy efficiency of these approaches in data-scarce regimes. This paper seeks to conduct a comprehensive empirical study that examines both model performance and energy efficiency of custom data augmentations and automated data augmentation selection strategies when combined with a lightweight object detector. The methods are evaluated in three different benchmark datasets in terms of their performance and energy consumption, and the Efficiency Factor is employed to gain insights into their effectiveness considering both performance and efficiency. Consequently, it is shown that in many cases, the performance gains of data augmentation strategies are overshadowed by their increased energy usage, necessitating the development of more energy efficient data augmentation strategies to address data scarcity.
\end{abstract}


\section{Introduction}
\label{sec:Introduction}
% V0.1
% In today's commercial and industrial ecosystem, Artificial Intelligence (AI) has become a key driver in upgrading services, products, and operations. This stems from the fact that AI has become the main driver of decision support, as its powerful nature supports real-time and high-fidelity predictions. 

% However, Machine Learning (ML) and Deep Learning (DL) model training can be an arduous task. In particular, AI model training is typically characterized by high computational costs due to the lengthy training sessions required to adapt to the large volumes of data \cite{marcus2018deep}. However, there are various cases where collecting abundant data can be challenging due to increased costs (e.g., industrial sector) or privacy concerns (e.g., healthcare sector). To address this data scarcity that is often encountered in these domains, various Data Augmentation (DA) approaches have been developed in recent years. These approaches can increase the dataset size effectively by applying various transformations to the data, which facilitates training and enhances model generalization. DAs can also improve model robustness by providing additional perspectives on identifiable data. At the same time, overlapping of the transformations, which can lead to model overfitting, can be easily controlled by configuring the hyperparameters and magnitude of the augmentations.

% V0.2
In today's commercial and industrial ecosystem, Artificial Intelligence (AI) has become a key driver in upgrading services, products, and operations, due to its powerful nature that supports real-time and high-fidelity predictions. However, training Machine Learning (ML) and Deep Learning (DL) models can be an arduous and computationally expensive task due to the lengthy training sessions required to adapt to large volumes of data \cite{marcus2018deep}. Additionally, there are various cases where collecting abundant data can be challenging due to increased costs (e.g., industrial sector) or privacy concerns (e.g., healthcare sector). To address this data scarcity that is often encountered in these domains, various Data Augmentation (DA) approaches have been developed in recent years. These approaches effectively increase dataset size by applying various transformations to the data, which facilitates training, enhances model generalization, and improves model robustness.

% V0.1
% Another issue with conventional AI models is that they fail to adapt to novel categories (classes) added after the model has been trained. This is a very common case in the constantly evolving industrial domain. However, for AI models to adapt to these new classes, they need to be retrained from scratch with the updated data, which is highly inefficient in terms of cost and energy consumption.

% V0.2
Another pervasive issue with conventional AI models in industrial applications is their struggle to adapt to novel categories (classes) added after training, requiring costly and energy-intensive retraining. While continual/incremental learning approaches can circumvent this limitation, they often come with added complexity, making them impractical for many industrial use cases.

% V0.1
% Given the pressing nature of dealing with the modern energy crisis, tackling these common limitations of AI systems is crucial \cite{schwartz2020green}. One of the latest and most innovative approaches to solving the aforementioned problems is Low-Shot Learning (LSL) and Few-Shot Learning (FSL). These techniques approach AI model training by leveraging prior knowledge and transferring it to novel tasks, requiring only small amounts of labeled data to generalize to these tasks and make accurate predictions. This significantly reduces the need for large datasets and extensive computational resources, making model training more efficient and effective, especially in resource-constrained environments. Furthermore, LSL/FSL enables models to adapt to new task requirements (e.g., additional classes) using only a small amount of data.

% V0.2
Given the pressing nature of dealing with the modern energy crisis, tackling these common limitations of AI systems is crucial \cite{schwartz2020green}. This is particularly true for real-time industrial applications where ML/DL models are expected to run on resource-constrained edge devices, rendering energy efficiency a critical requirement. Recently, low-shot learning (LSL) and few-shot learning (FSL) have emerged as promising solutions that leverage prior knowledge and transfer it to novel tasks, requiring only small amounts of labeled data to generalize and make accurate predictions. This significantly reduces the need for large datasets and extensive computational resources, making model training more efficient and effective while enabling adaptation to new task requirements (e.g., additional classes) using only a small amount of data.

% V0.1
% This work investigates the overall efficiency of various DA approaches used during the finetuning of a lightweight object detector to novel downstream under data scarcity (LSL and FSL scenarios). Specifically, the main objective is to provide a clear perspective of the efficacy of these techniques in low-capacity systems, like edge devices and embedded systems, by evaluating both their energy efficiency during the finetuning process as well as the generalization capabilities of the resulting models. The presented empirical study rigorously evaluates a set of different DA methods under varying levels of data availability and provides an in-depth analysis of the results. The contributions of this research can be summarized as follows:

% V0.2
Despite the emergence of various energy-efficient LSL/FSL approaches in recent years, limited attention has been given to evaluating the energy efficiency of DA methods for LSL/FSL, which are widely used in practice. This work aims to fill this gap by investigating the overall efficiency of such DA approaches used during finetuning a lightweight object detector to novel downstream tasks under data scarcity (LSL and FSL scenarios). Specifically, the main objective of this empirical study is to provide a clear perspective on the efficacy of these techniques in low-capacity systems, like edge devices and embedded systems, by evaluating both their energy efficiency during the finetuning process as well as the generalization capabilities of the resulting models under varying levels of data availability. The contributions of this research can be summarized as follows:

\begin{itemize}
    
    \item A performance and energy efficiency analysis of utilizing various DA strategies during finetuning on three benchmark datasets for low- and few-shot object detection is provided.

    \item An ablation study further examining the performance differences between custom and automated DA methods is presented.

    \item An evaluation of DA methods in the context of LSL/FSL using a customized Efficiency Factor metric is conducted for the fist time.

\end{itemize}

% The rest of this paper is organised as follows. Section \ref{sec:Related_Work} presents the related work while Section \ref{sec:Methodology} showcases the employed methodology. Section \ref{sec:Experimental_Setting} delineates the experimental configuration of this work the results of which are presented in Sections \ref{FSL} and \ref{LSL}. Finally, \ref{sec:Conclusion} concludes this work.

\section{Related Work}
\label{sec:Related_Work}
\subsection{Low/Few-Shot Object Detection}

Following the broader paradigm of LSL and FSL, low/few-shot detection techniques can be categorized into meta-learning and finetuning-based approaches. Meta-learning approaches involve creating low/few-shot tasks during training and learning how to transfer knowledge to novel tasks in a class-agnostic manner. Several existing object detector architectures have been adapted for this purpose, including Meta Faster R-CNN \cite{han2022meta} and Meta-DETR \cite{zhang2022meta}. On the other hand, finetuning-based approaches are pretrained in a base dataset with abundant annotated data and then finetuned in the low/few-shot tasks incorporating techniques such as finetuning only the final classification layer \cite{wang2020frustratingly}, contrastive learning \cite{sun2021fsce} and gradient scaling and stopping \cite{qiao2021defrcn}.

In recent years, new approaches have been developed to address data scarcity in the LSL and FLS settings using DAs. These include employing a hallucinator network to generate novel image \cite{wang2018low} and RoI \cite{zhang2021hallucination} samples, using a Variational Autoencoder (VAE) that generates novel features in the model's latent space \cite{xu2023generating}, processing RoIs extracted from the original images in multiple scales \cite{wu2020multi}, and semantically separating foreground objects from their backgrounds and fusing them with novel ones \cite{wang2024snida}. However, it remains unclear whether these approaches are optimal when considering energy efficiency due to their increased complexity.

\subsection{Energy Efficient Object Detection}

Green AI has recently emerged as a nascent area that aims to address the energy efficiency and carbon footprint concerns of modern deep learning approaches by developing more energy efficient models. In the field of computer vision, in \cite{yang2019ecc}, model energy consumption is estimated using a bilinear regression model and used in a constrained optimization problem to obtain a compressed and energy efficient model. Additionally, in \cite{stamoulis2018designing} the architecture settings of a CNN are treated as hyperparameters to be optimized using Bayesian optimization, with the optimization problem taking into consideration both model performance and energy efficiency. 

Recently, energy efficiency has also been examined in the context of object detection in \cite{tu2023femtodet}, which evaluates different components of object detectors regarding energy efficiency and proposes an architecture and data augmentation strategy based on their findings. Nevertheless, there has been limited assessment of the energy efficiency of LSL/FSL for object detection, with the exception of \cite{tsoumplekas2024evaluating}, where the authors assess the efficiency of finetuning-based approaches in this context and propose a novel metric aimed at consolidating model performance and energy efficiency into a single value.

\subsection{Data Augmentation}
DAs for deep learning-based methods, particularly for computer vision applications, have been extensively researched in previous years. In the context of object detection, DAs usually refer to basic hand-crafted operations selected for specific models and datasets and can be broadly categorized as image manipulation (e.g., rotation, translation, shearing), image erasing (e.g., random erasing, Cutout), and image mixing (e.g., mixup \cite{zhang2018mixup}). More recently, a novel line of DA approaches has also emerged where selecting a DA strategy is formulated as a discrete space search problem and is solved using reinforcement learning \cite{cubuk2019autoaugment}, random selection \cite{cubuk2020randaugment} or by enforcing consistency between original and augmented images \cite{hendrycks2019augmix}. However, the assessment of these techniques in terms of energy efficiency has been limited, especially in the context of LSL and FSL. Our study examines both hand-crafted DA techniques and automated methods for their performance vs energy efficiency trade-offs.

\section{Methodology}
\label{sec:Methodology}
\subsection{Problem Formulation}

In the context of few-shot object detection (FSOD) and low-shot object detection (LSOD), the goal is to develop object detectors capable of rapidly adapting to novel downstream tasks containing only a few training images of previously unseen objects. Following the standard formulation introduced in \cite{kang2019few}, an object detector is initially trained on a \textit{base} dataset with abundant data of $\mathcal{C}_{base}$ classes and is subsequently adapted to a \textit{novel} dataset with limited data of $\mathcal{C}_{novel}$ previously unseen classes, i.e., $\mathcal{C}_{base} \cap \mathcal{C}_{novel} = \emptyset$. More specifically, we define the \textit{base} dataset as $\mathcal{D}_{base} = \{(I_n, y_n)\}_{n=1}^N$, where $I_n \in \mathcal{I}$ is an input image and $y_n \in \mathcal{Y}$ is its corresponding label and bounding box annotations. In particular, image $I_n$ contains $B_n$ objects, so $y_n = \{ (c_b, box_b) \}_{b=1}^{B_n}$, where $c_i \in \mathcal{C}_{base}$ is the object label and $box_b = (x_b,y_b,w_b,h_b)$ is the object's box location. Typically, $N$ is large, and as a result, an object detector $f_{\theta}$, parameterized by $\theta$, can be effectively trained in this dataset using standard supervised learning approaches. 

Following the pretraining of the model on $\mathcal{D}_{base}$, the next step is to adapt $f_{\theta}$ in the \textit{novel} dataset which can be defined as $\mathcal{D}_{novel} = \{ (I_m, y_m) \}_{m=1}^M$, where $I_m$ is the input image but now $y_m=\{(c_b,box_b) \}_{b=1}^{B_m}$ with $c_b \in \mathcal{C}_{novel}$. In $\mathcal{D}_{novel}$ there are only $K$ labeled bounding boxes for each class, and the images having these bounding boxes constitute the support set $\mathcal{S}$ of $\mathcal{D}_{novel}$. Additionally, $\mathcal{D}_{novel}$ also contains a set of images used for evaluating the final model's performance in the $\mathcal{C}_{novel}$ classes, called the query set $\mathcal{Q}$. 

Based on this formulation, a straightforward way to adapt $f_{\theta}$ on $\mathcal{D}_{novel}$ is to finetune the model on $\mathcal{S}$ replacing only its final classification layer. Finetuning can be either full (whole model) or partial (e.g., frozen backbone), and the resulting model $f_{\theta'}$ is finally evaluated on $\mathcal{Q}$.

\subsection{Real-Time Detection with YOLO}
This work aims to evaluate the performance and energy efficiency of DA strategies used for LSL/FSL. One effective way to demonstrate this is by testing it on computation-intensive tasks, such as object detection. For this study, the YOLOv8 architecture is utilized, as it is the most widely used real-time object detector at the time of this work. Compared to older versions like YOLOv3 and YOLOv5, this version demonstrates the best balance between efficiency and accuracy.

The YOLOv8 model consists of two main modules: the backbone feature extractor and the prediction head. The former is based on a customized version of the CSPDarknet53 architecture, selected for its highly generalizable extracted feature representations. The structure of this network is based on a feature pyramid network (FPN), enabling the identification of objects of varying sizes and scales within an image by extracting characteristics at different levels. The prediction head is a convolutional network followed by three different detection modules whose inputs are features extracted from different levels of the FPN, allowing for multi-scaled object detection. These modules are designed to ensure that the model accurately predicts item positions and classes across a wide range of data with minimal overhead, making YOLOv8 particularly effective for real-time object detection tasks.

\subsection{Data Augmentation Strategies}

In this study, we examine two main lines of work: custom DA strategies, where the model developer defines the DA operations, and automated DA selection strategies, where the DA operations are selected using a search algorithm.

For the custom DA strategies, we specifically address the challenges of working with limited data (LSL and FSL settings) and the risk of overfitting the small training datasets in these cases. In the first set of augmentations, we generate novel images by applying the Sharpen, Solarize, and Superpixel operations separately to each original image, increasing the available training images by a factor of four. In the second set of custom DAs, we aim to mitigate overfitting by increasing the diversity among the available images by randomly selecting a subset of images and separately applying Rotation, Translation, Scaling, Shearing, Perspective changing, Flipping, HSV manipulation, and Mosaic operations with specific probabilities.

As for the automated DA selection methods, AutoAugment \cite{cubuk2019autoaugment}, RandAugment \cite{cubuk2020randaugment}, and AugMix \cite{hendrycks2019augmix} are examined. In AutoAugment \cite{cubuk2019autoaugment}, selecting the optimal DA methods to be applied is formulated as a discrete search problem. Given a pool of data augmentation methods, an LSTM-based controller model is utilized to select DAs as well as their magnitude and probability of being applied. These DAs are then used to train a target neural network for a specific task, and the model's performance is used as a reward to optimize the LSTM controller using Reinforcement Learning.

To simplify the training process and reduce computational costs of AutoAugment, RandAugment \cite{cubuk2020randaugment} randomly selects $N$ DA methods from the DAs pool and applies them to each image with a magnitude of $M$, minimizing the method's hyperparameters to just two.

Finally, in AugMix \cite{hendrycks2019augmix} augmented images $I_{aug}$ are obtained as the weighted sum of images created using $K$ randomly selected chains of DAs applied on the original image $I_{orig}$, i.e., $I_{aug} = \sum_{k=1}^K w_k chain_k(I_{orig})$, where weights $w_k$ are randomly sampled from a Dirichlet distribution and $chain_k$ is a randomly selected sequence of DA operations. To ensure consistency between the original and the augmented images, the final augmented images are obtained as a convex combination of $I_{orig}$ and $I_{aug}$, specifically $I_{augmix} = mI_{orig} + (1-m)I_{aug}$, where $m \in (0, 1)$ is the mixing coefficient. To further ensure consistency between the representations of the original and augmented images, due to their semantic similarity, the Jensen-Shannon divergence among the posterior distributions of the original and augmented images is minimized as part of the model's objective function.

\section{Experimental Setting}
\label{sec:Experimental_Setting}
\subsection{Datasets}

In the following experiments, three datasets containing images of objects and hazards commonly found in industrial settings are used to assess the examined DA techniques. Table \ref{dataset_details} summarizes the main details of these three datasets. For each dataset, the images in the training set are utilized to create the low/few-shot training tasks for finetuning the object detectors. Model performance on the validation set images is used to determine the number of training epochs through an early stopping procedure. Finally, all reported performance metrics of the finetuned models are based on evaluation using the test set images.

\begin{table}[h]
\centering
\caption{Dataset Characteristics}
\label{dataset_details}
\resizebox{\columnwidth}{!}{%
\begin{tabular}{llll}
\toprule
\textbf{Dataset} & \textbf{Image Type} & \textbf{Classes} & \textbf{Train/Val/Test Split} \\
\midrule
PPE \cite{sesis2022robust} & Personal protective equipment (PPE) used by firefighters & 4 & 280/34/31 \\
Fire & Scenes of fires & 1 & 2768/124/698 \\
CS \cite{worker-safety_dataset} & PPE used by workers in construction sites & 5 & 997/119/90 \\
\bottomrule
\end{tabular}
}
\end{table}

\subsection{Model Settings}

In the following experiments, we utilize the YOLOv8n variant of the YOLOv8 model, pretrained on the MS COCO dataset, as our initial model before finetuning. The model contains approximately 3.2M parameters, but only the parameters of the three detection modules are finetuned, resulting in $\approx$750K trainable parameters during finetuning. The initial model is finetuned for 1000 epochs using Early Stopping with patience set to 100 epochs. During optimization, AdamW with a learning rate of 0.01 is employed, and the batch size is set to 32. The resulting model is denoted as \texttt{FT}.

% V0.1
% As for the DA strategies followed, all images are initially resized to $640\times640$ pixels. All custom augmentation techniques are implemented using the Albumentations \cite{buslaev2020albumentations} library with default settings provided. We denote augmentations targeting data scarcity as \texttt{DA(1)}, augmentations tackling overfitting as \texttt{DA(2)}, and their combination as \texttt{DA(1+2)}. As for the automated DA selection techniques, the pool of DAs includes translation, scaling, horizontal flipping, hue, saturation, and brightness, following the default settings provided by the Ultralytics \cite{Jocher_Ultralytics_YOLO_2023} library.

% V0.2
As for the DA strategies followed, all images are initially resized to $640\times640$ pixels. All custom augmentation techniques are implemented using the Albumentations \cite{buslaev2020albumentations} library with default settings provided. We denote augmentations targeting data scarcity as \texttt{DA(1)}, augmentations tackling overfitting as \texttt{DA(2)}, and their combination as \texttt{DA(1+2)}. As for the automated DA selection techniques, the pool of DAs includes translation, scaling, horizontal flipping, hue, saturation, and brightness.

Finally, all experiments were conducted on a NVIDIA GeForce RTX 2080 Ti using a 64 GB RAM.

\subsection{Evaluation Metrics}

% V0.1
% Following the standard evaluation procedures for object detection models, we use $AP_{50}$, which represents the model's Average Precision (AP) using a fixed Intersection over Union (IoU) threshold of 50\%, to evaluate model performance. In addition to model performance, we are also interested in assessing the energy efficiency of the models during the finetuning process. This is achieved by measuring each model's energy consumed during finetuning using the CodeCarbon library. Finally, to consider both model performance and efficiency, we utilize a modified form of the Efficiency Factor ($EF$) metric, which was introduced in \cite{tsoumplekas2024evaluating} and is defined as follows:

% V0.2
Following the standard evaluation procedures for object detection models, we use $AP_{50}$, which represents the model's Average Precision (AP) using a fixed Intersection over Union (IoU) threshold of 50\%, to evaluate model performance. In addition to model performance, we are also interested in assessing the energy efficiency of the models during the finetuning process. This is achieved by measuring each model's energy consumed during finetuning using the CodeCarbon library. While the specific hardware configuration can impact these measurements, the relative differences between the various examined approaches will remain consistent, making the results valuable even to users with different hardware setups. Finally, to consider both model performance and efficiency, we utilize a modified form of the Efficiency Factor ($EF$) metric, which was introduced in \cite{tsoumplekas2024evaluating} and is defined as follows:

\begin{equation} \label{ef_metric}
    EF = \frac{AP_{50}}{1+EC}
\end{equation}

\noindent where $AP_{50} \in [0, 100]$ is used instead of the $mAP$ in the original formulation, and $EC \in (0, +\infty)$ is the model’s energy consumption measured in $Wh$.

\section{Few-Shot Learning Results} \label{FSL}
In the FSL setting, the models are finetuned using $N-way$ $K-shot$ tasks, where $N$ is the number of classes in the dataset and $K \in \{1,2,3,5,10,30\}$ is the number of bounding box annotated objects used as support set samples for each class, while the entire test set of each dataset is used as the query set.

\begin{table}[]
\centering
\caption{Test set $AP_{50}$ for different training strategies and number of shots in the FSL scenario.}
\label{fsl_map}
\resizebox{0.9\columnwidth}{!}{%
\begin{tabular}{llrrrrrr}
\toprule
\multicolumn{1}{c}{\multirow{2}{*}{\textbf{Dataset}}} & \multicolumn{1}{c}{\multirow{2}{*}{\textbf{Model}}} & \multicolumn{5}{c}{\textbf{Shots}}                                      \\ \cmidrule(l){3-8}
        & & \textbf{1} & \textbf{2} & \textbf{3} & \textbf{5} & \textbf{10} & \textbf{30} \\
\midrule
\multicolumn{1}{l}{\multirow{7}{*}{\textbf{PPE}}}
& FT & 12.23 & 10.24 & 12.42 & 12.00 & 17.70 & 23.99  \\
& FT+DA(1) & 11.80 & 10.51 & 12.79 & 12.60 & 17.69 & 23.17 \\
& FT+DA(2) & \textbf{12.96} & \textbf{11.12} & 14.15 & 13.77 & \textbf{25.89} & 27.47 \\
& FT+DA(1+2) & 12.63 & 11.07 & \textbf{15.71} & 15.16 & 25.56 & 28.11  \\
& FT+AutoAugment\cite{cubuk2019autoaugment} & 12.17 & 10.41 & 15.11 & \textbf{15.71} & 23.22 & \textbf{33.59} \\
& FT+RandAugment\cite{cubuk2020randaugment} & 12.17 & 10.41 & 15.11 & \textbf{15.71} & 23.22 & \textbf{33.59} \\
& FT+AugMix\cite{hendrycks2019augmix} & 12.17 & 10.41 & 15.11 & \textbf{15.71} & 23.22 & \textbf{33.59} \\
\midrule
\multicolumn{1}{l}{\multirow{7}{*}{\textbf{Fire}}} 
& FT &  1.54 & 1.84 & \textbf{2.05} & 1.84 & 2.10 & 3.02  \\
& FT+DA(1) & \textbf{2.34} & \textbf{1.91} & 1.82 & \textbf{1.98} & 2.18 & 2.64 \\
& FT+DA(2) & 1.80 & 1.38 & 0.80 & 1.59 & \textbf{2.20} & 2.01 \\
& FT+DA(1+2) &  2.08 & 1.68 & 1.32 & 0.63 & 1.95 & 3.32  \\
& FT+AutoAugment\cite{cubuk2019autoaugment} & 1.82 & 1.59 & 1.66 & 1.60 & 1.97 & \textbf{3.53} \\
& FT+RandAugment\cite{cubuk2020randaugment} & 1.82 & 1.59 & 1.66 & 1.60 & 1.97 & \textbf{3.53} \\
& FT+AugMix\cite{hendrycks2019augmix} & 1.82 & 1.59 & 1.66 & 1.60 & 1.97 & \textbf{3.53} \\
\midrule
\multicolumn{1}{l}{\multirow{7}{*}{\textbf{CS}}} 
& FT & 11.58 & 14.51 & 12.99 & 14.06 & 18.49 & 32.25  \\
& FT+DA(1) & 11.67 & 14.28 & 12.72 & 14.49 & 17.19 & 25.17 \\
& FT+DA(2) & 11.78 & \textbf{16.29} & \textbf{14.50} & 19.14 & 26.96 & 37.84 \\
& FT+DA(1+2) &  10.31 & 14.30 & 13.30 & \textbf{19.34} & 27.97 & 36.00  \\
& FT+AutoAugment\cite{cubuk2019autoaugment} & \textbf{12.07} & 15.27 & 14.15 & 15.18 & \textbf{33.33} & \textbf{45.68} \\
& FT+RandAugment\cite{cubuk2020randaugment} & \textbf{12.07} & 15.27 & 14.15 & 15.18 & \textbf{33.33} & \textbf{45.68} \\
& FT+AugMix\cite{hendrycks2019augmix} & \textbf{12.07} & 15.27 & 14.15 & 15.18 & \textbf{33.33} & \textbf{45.68} \\
\bottomrule
\end{tabular}
}
\end{table}

\subsection{Main Results}

Table \ref{fsl_map} shows the model performance of different DA strategies for each dataset's support set for varying shots. It is evident that increasing the number of shots in all cases leads to improved performance since more training samples become available during the finetuning process. Additionally, the three examined automated DA selection methods (\texttt{FT+AutoAugment}, \texttt{FT+RandAugment}, \texttt{FT+AugMix}) demonstrate the same performance in all cases, possibly due to the limited DA pool that leads to the selection of the same DAs by all three methods.

For the PPE dataset, model performance is similar for a small number of shots. However, the gap increases for a greater number of shots, with automated DA selection methods leading to improved results (33.59\%) compared to custom DAs (28.11\% for \texttt{FT+DA(1+2)}) and finetuning without DA (23.99\%). Similar conclusions can be drawn for the CS dataset where automated DA selection methods significantly outperform custom DA (45.68\% vs. 37.84\% for \texttt{FT+DA(2)}) and vanilla \texttt{FT} (45.68\% vs 32.25\%). In the Fire dataset, custom DA methods and mostly \texttt{FT+DA(1)} produce strong results for a small number of shots, which could be attributed to the fact that it generates additional samples and assists in tackling the extreme data scarcity in these settings. However, in the 30-shot scenario, automated DA selection methods lead to the best results. Finally, it is worth noticing that since the number of classes affects the number of training samples available in the $N-way$ $K-shot$ formulation of tasks in FSL, models struggle in datasets such as Fire with only one class available, and their performance improves in datasets with more classes such as PPE and CS.

\begin{table}[]
\centering
\caption{Total energy consumption in Wh during training for different training strategies and number of shots in the FSL scenario.}
\label{fsl_energy}
\resizebox{0.9\columnwidth}{!}{%
\begin{tabular}{llrrrrrr}
\toprule
\multicolumn{1}{l}{\multirow{2}{*}{\textbf{Dataset}}} & \multicolumn{1}{l}{\multirow{2}{*}{\textbf{Model}}} & \multicolumn{5}{c}{\textbf{Shots}}                                      \\ \cmidrule(l){3-8}
        & & \multicolumn{1}{c}{\textbf{1}} & \multicolumn{1}{c}{\textbf{2}} & \multicolumn{1}{c}{\textbf{3}} & \multicolumn{1}{c}{\textbf{5}} & \multicolumn{1}{c}{\textbf{10}} & \multicolumn{1}{c}{\textbf{30}} \\
\midrule
\multicolumn{1}{l}{\multirow{7}{*}{\textbf{PPE}}}
& FT & \textbf{3.787} & \textbf{3.927} & \textbf{4.406} & \textbf{4.942} & \textbf{7.001} & 17.003 \\
& FT+DA(1) & 4.619 & 4.673 & 5.111 & 5.758 & 11.618 & 29.394 \\
& FT+DA(2) & 5.835 & 9.603 & 12.148 & 18.297 & 53.684 & 231.66 \\
& FT+DA(1+2) & 12.281 & 46.428 & 60.022 & 128.461 & 244.561 & 685.058 \\
& FT+AutoAugment\cite{cubuk2019autoaugment} & 4.255 & 6.460 & 9.161 & 8.111 & 9.208 & 11.175 \\
& FT+RandAugment\cite{cubuk2020randaugment} & 4.311 & 6.425 & 9.323 & 8.229 & 9.104 & \textbf{10.836} \\
& FT+AugMix\cite{hendrycks2019augmix} & 4.231 & 6.311 & 9.224 & 8.003 & 9.100 & 10.901 \\
\midrule
\multicolumn{1}{l}{\multirow{7}{*}{\textbf{Fire}}} 
& FT & 9.581 & 8.932 & \textbf{8.324} & 11.079 & 9.297 & \textbf{8.276} \\
& FT+DA(1) & \textbf{8.365} & \textbf{7.222} & 8.734 & \textbf{7.611} & \textbf{9.198} & 9.523 \\
& FT+DA(2) & 9.760 & 9.591 & 12.701 & 9.352 & 18.952 & 34.822 \\
& FT+DA(1+2) & 8.846 & 13.256 & 14.571 & 18.389 & 47.468 & 114.692 \\
& FT+AutoAugment\cite{cubuk2019autoaugment} & 10.636 & 10.786 & 10.424 & 9.638 & 12.718 & 11.851 \\
& FT+RandAugment\cite{cubuk2020randaugment} & 10.760 & 10.530 & 10.424 & 9.716 & 12.440 & 11.729 \\
& FT+AugMix\cite{hendrycks2019augmix} & 10.480 & 10.549 & 10.629 & 9.528 & 12.661 & 11.758 \\
\midrule
\multicolumn{1}{l}{\multirow{7}{*}{\textbf{CS}}}
& FT & \textbf{6.398} & \textbf{6.878} & \textbf{6.998} & \textbf{7.122} & 13.988 & 16.969 \\
& FT+DA(1) & 6.865 & 7.257 & 7.417 & 8.515 & \textbf{11.913} & 42.546 \\
& FT+DA(2) & 7.934 & 10.181 & 12.309 & 24.966 & 74.763 & 177.151 \\
& FT+DA(1+2) & 15.528 & 37.117 & 58.231 & 146.523 & 247.867 & 654.121 \\
& FT+AutoAugment\cite{cubuk2019autoaugment} & 6.444 & 7.092 & 7.112 & 7.442 & 15.119 & 16.027 \\
& FT+RandAugment\cite{cubuk2020randaugment} & 6.462 & 7.136 & 7.172 & 7.575 & 14.942 & 15.970 \\
& FT+AugMix\cite{hendrycks2019augmix} & 6.431 & 7.053 & 6.988 & 7.433 & 14.946 & \textbf{15.831} \\
\bottomrule
\end{tabular}
}
\end{table}

Table \ref{fsl_energy} includes the energy consumed by each examined model during the finetuning process for each dataset and a varying number of shots. Generally, increasing the number of shots in the PPE and CS datasets leads to increased energy consumption. This behavior is expected since increasing the shots leads to more samples that need to be processed and, consequently, more finetuning epochs for the models to converge. Only a few minor exceptions emerge, such as for the automated DA selection methods in the 5- and 10-shot cases in the PPE dataset and \texttt{FT+AugMix} in the 3-shot scenario of the CS dataset, due to the early stopping mechanism employed. On the other hand, this relationship between the energy consumption and the number of shots is unclear in the Fire dataset, where training with fewer samples might lead to increased energy consumption. However, this instability could be attributed to the extremely small number of samples available in each scenario, which is highly affected by the randomness introduced in the sampling process when formulating the training tasks, thus affecting the convergence of the models. It is worth noticing that in the PPE and CS datasets, including DA methods in the finetuning process leads to increased energy consumption. However, automated DA selection approaches, such as \texttt{FT+RandAugment} for the PPE and \texttt{FT+AugMix} for the CS, can be beneficial in the 30-shot scenario since they could allow for a faster convergence compared to vanilla \texttt{FT}. For the Fire dataset, where data is extremely scarce, using a DA strategy that increases the number of available training samples such as \texttt{FT+DA(1)} can lead to faster convergence and reduced energy consumption compared to other DA strategies. Finally, it is worth noting that even though \texttt{FT+DA(1+2)} combines the DA methods used in \texttt{FT+DA(1)} and \texttt{FT+DA(2)}, its energy consumption is significantly larger than the sum of the energy consumed when applying these two methods separately.

\begin{table}[]
\centering
\caption{Efficiency factor (EF) metric values for different training strategies and number of shots in the FSL scenario.}
\label{fsl_ef}
\resizebox{0.9\columnwidth}{!}{%
\begin{tabular}{llrrrrrr}
\toprule
\multicolumn{1}{l}{\multirow{2}{*}{\textbf{Dataset}}} & \multicolumn{1}{l}{\multirow{2}{*}{\textbf{Model}}} & \multicolumn{5}{c}{\textbf{Shots}}                                      \\ \cmidrule(l){3-8}
        & & \multicolumn{1}{c}{\textbf{1}} & \multicolumn{1}{c}{\textbf{2}} & \multicolumn{1}{c}{\textbf{3}} & \multicolumn{1}{c}{\textbf{5}} & \multicolumn{1}{c}{\textbf{10}} & \multicolumn{1}{c}{\textbf{30}} \\
\midrule
\multicolumn{1}{l}{\multirow{7}{*}{\textbf{PPE}}}
& FT & \textbf{2.506} & \textbf{2.092} & \textbf{2.316} & \textbf{2.067} & \textbf{2.434} & 1.554 \\
& FT+DA(1) & 2.072 & 1.881 & 2.123 & 1.866 & 1.426 & 0.779 \\
& FT+DA(2) & 1.872 & 1.111 & 1.109 & 0.758 & 0.472 & 0.129 \\
& FT+DA(1+2) & 0.940 & 0.281 & 0.259 & 0.121 & 0.104 & 0.043 \\
& FT+AutoAugment\cite{cubuk2019autoaugment} & 2.282 & 1.538 & 1.784 & 1.916 & 2.342 & 2.813 \\
& FT+RandAugment\cite{cubuk2020randaugment} & 2.261 & 1.541 & 1.758 & 1.927 & 2.329 & \textbf{2.890} \\
& FT+AugMix\cite{hendrycks2019augmix} & 2.300 & 1.570 & 1.782 & 1.935 & 2.344 & 2.879 \\
\midrule
\multicolumn{1}{l}{\multirow{7}{*}{\textbf{Fire}}} 
& FT & 0.159 & 0.190 & \textbf{0.226} & 0.153 & 0.207 & \textbf{0.324} \\
& FT+DA(1) & \textbf{0.247} & \textbf{0.233} & 0.197 & \textbf{0.230} & \textbf{0.220} & 0.251 \\
& FT+DA(2) & 0.166 & 0.133 & 0.061 & 0.167 & 0.130 & 0.058 \\
& FT+DA(1+2) & 0.212 & 0.144 & 0.093 & 0.034 & 0.041 & 0.029 \\
& FT+AutoAugment\cite{cubuk2019autoaugment} & 0.174 & 0.134 & 0.150 & 0.150 & 0.148 & 0.300 \\
& FT+RandAugment\cite{cubuk2020randaugment} & 0.173 & 0.138 & 0.150 & 0.149 & 0.151 & 0.300 \\
& FT+AugMix\cite{hendrycks2019augmix} & 0.174 & 0.137 & 0.147 & 0.151 & 0.149 & 0.301 \\
\midrule
\multicolumn{1}{l}{\multirow{7}{*}{\textbf{CS}}}
& FT & 1.574 & 1.848 & 1.622 & 1.730 & 1.466 & 1.807 \\
& FT+DA(1) & 1.489 & 1.736 & 1.510 & 1.524 & 1.316 & 0.625 \\
& FT+DA(2) & 1.324 & 1.458 & 1.094 & 0.726 & 0.370 & 0.213 \\
& FT+DA(1+2) & 0.623 & 0.379 & 0.226 & 0.130 & 0.113 & 0.055 \\
& FT+AutoAugment\cite{cubuk2019autoaugment} & \textbf{1.630} & 1.894 & 1.750 & 1.801 & 2.108 & 2.725 \\
& FT+RandAugment\cite{cubuk2020randaugment} & 1.622 & 1.879 & 1.742 & 1.776 & \textbf{2.131} & 2.735 \\
& FT+AugMix\cite{hendrycks2019augmix} & 1.629 & \textbf{1.902} & \textbf{1.778} & \textbf{1.804} & 2.115 & \textbf{2.749} \\
\bottomrule
\end{tabular}
}
\end{table}

Table \ref{fsl_ef} displays the $EF$ metric of each examined DA strategy for each number of shots in each dataset. Generally, $EF$ improves with higher $AP_{50}$ values and lower energy consumption. Consequently, we anticipate that models fulfilling either or both desiderata will achieve high $EF$ values. For the PPE dataset, \texttt{FT} achieves the best $EF$ values due to its low energy consumption while maintaining a solid performance in $AP_{50}$. In the Fire dataset, \texttt{FT+DA(1)} performs particularly well, followed by \texttt{FT}. It is worth noting that in these two datasets, the models with the lowest energy consumption also have the highest $EF$ values, highlighting the impact of energy consumption on the $EF$ metric. However, in the CS dataset, the best $EF$ values are achieved by the automated DA selection methods due to their significant performance boost in $AP_{50}$, despite \texttt{FT} and \texttt{FT+DA(1)} having the lowest energy consumption in most cases. Overall, as the number of shots increases, $EF$ values tend to decrease as energy consumption rises faster than $AP_{50}$. However, this trend does not hold for the automated DA selection techniques, as their $EF$ values are the highest in the 30-shot scenario for all three datasets.

\subsection{Effect of Custom Augmentations}

\begin{figure}[htp]
  \centering
  \subfigure[PPE Dataset]{\includegraphics[scale=0.26]{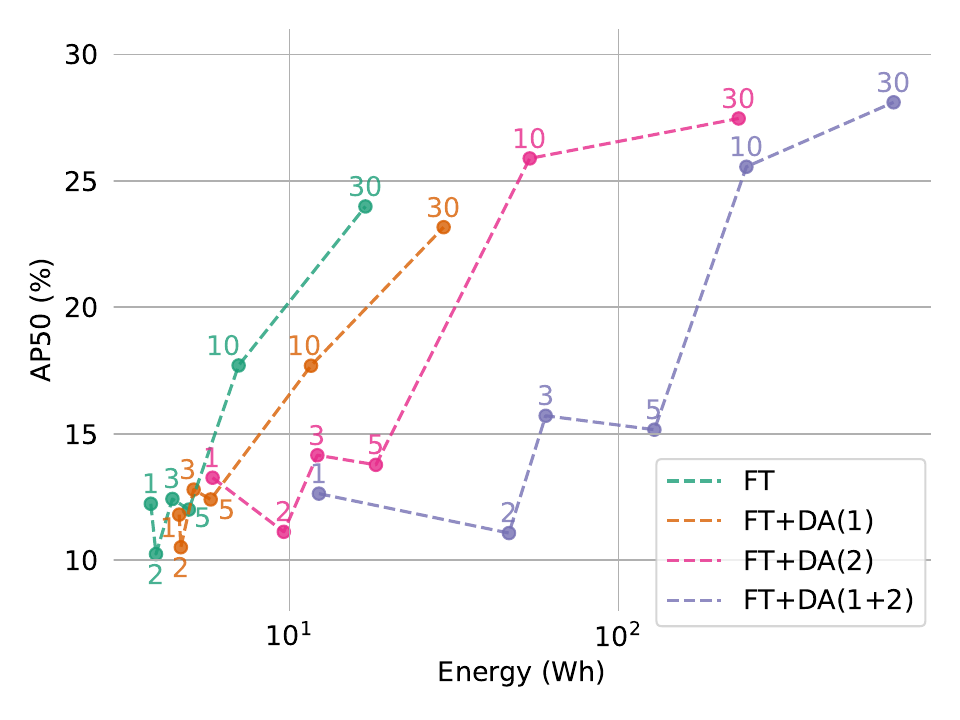}}
  \subfigure[CS Dataset]{\includegraphics[scale=0.26]{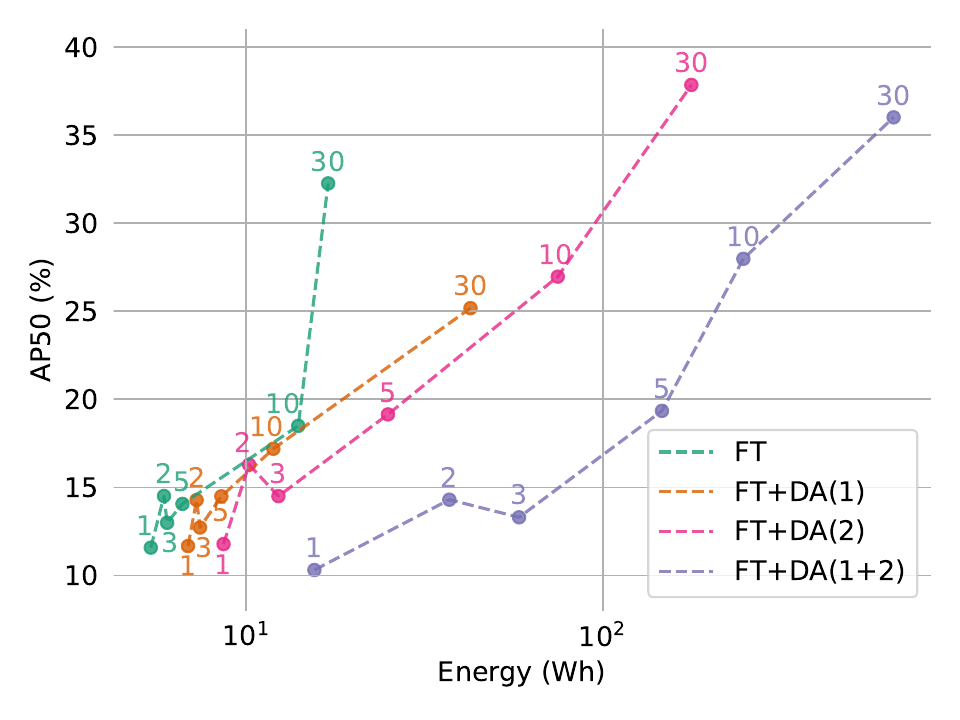}}
  \caption{$AP_{50}$ with respect to energy consumption for different DA
strategies and numbers of shots.}
  \label{energy_map_figure}
\end{figure}

Fig. \ref{energy_map_figure} illustrates model performance in terms of $AP_{50}$ with respect to the energy consumed during finetuning using vanilla \texttt{FT} or one of the three examined custom DA approaches. We have included the relevant plots only for the PPE and CS datasets since the extreme data scarcity in the Fire dataset does not allow for a valid interpretation of the obtained results. Overall, in both datasets, it is evident that \texttt{FT} leads to the most optimal performance vs. energy efficiency trade-off followed by \texttt{FT+DA(1)}, \texttt{FT+DA(2)}, and finally \texttt{FT+DA(1+2)}. It is also worth noticing that almost all curves corresponding to a different method (shown with different colors in the plot) can be approximated by a straight line, indicating that energy consumption grows exponentially compared to $AP_{50}$. This also verifies our findings that $EF$ values tend to drop as the number of shots increases. The only case where this does not seem to be true is for \texttt{FT} in the CS dataset, where there appears to be a linear relationship between model performance and energy consumption.
 
\subsection{Effect of automated augmentations}

\begin{figure*}[htp]
  \centering
  \subfigure[PPE Dataset]{\includegraphics[scale=0.33]{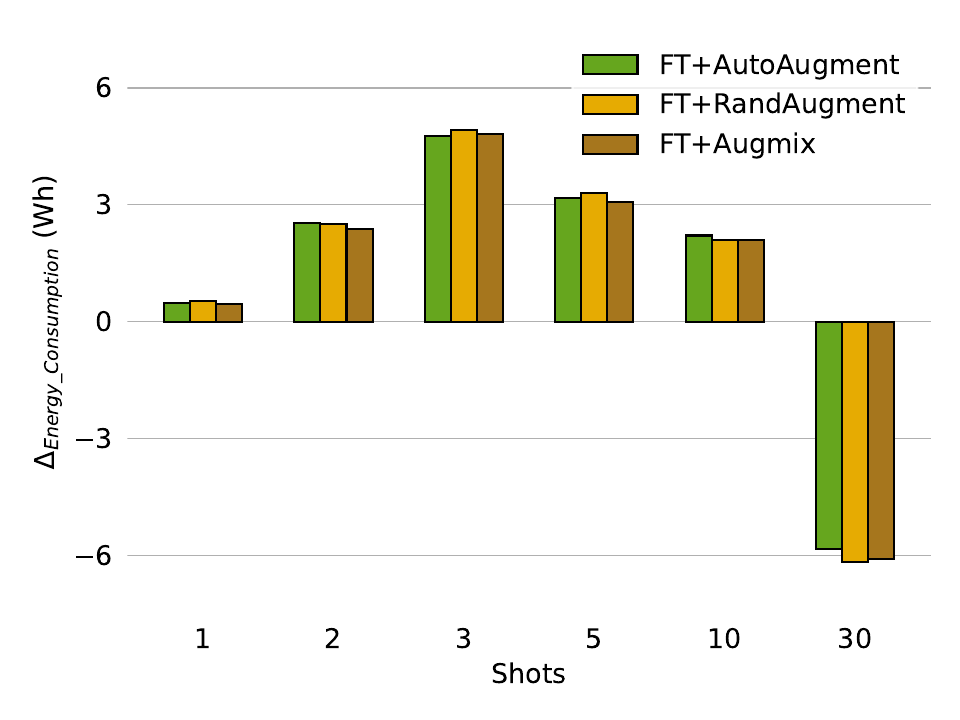}}
  \subfigure[Fire Dataset]{\includegraphics[scale=0.33]{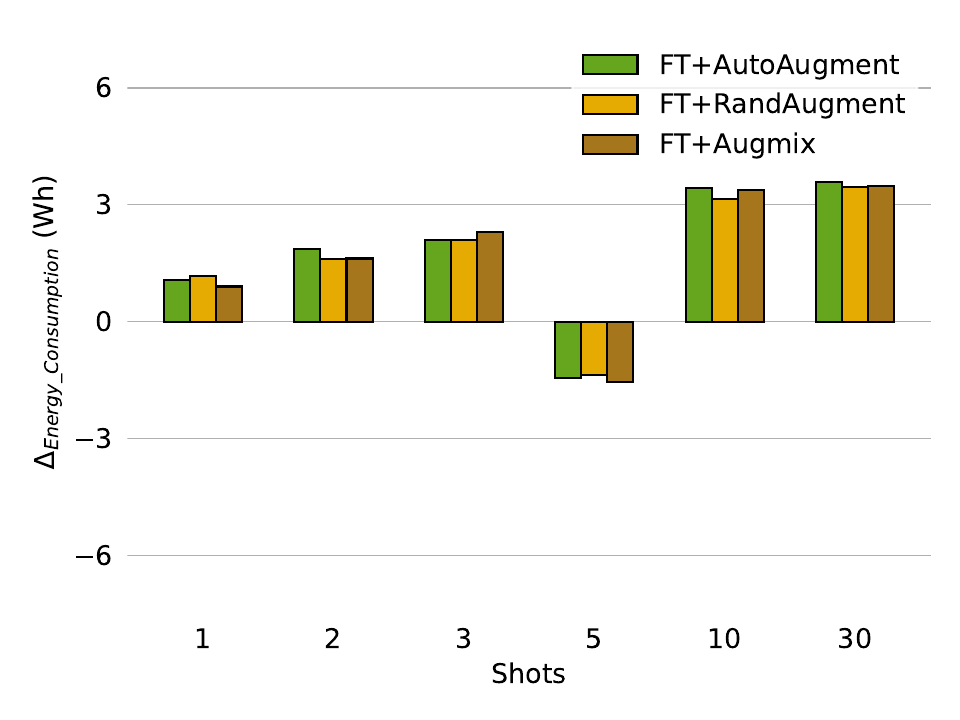}}
  \subfigure[CS Dataset]{\includegraphics[scale=0.33]{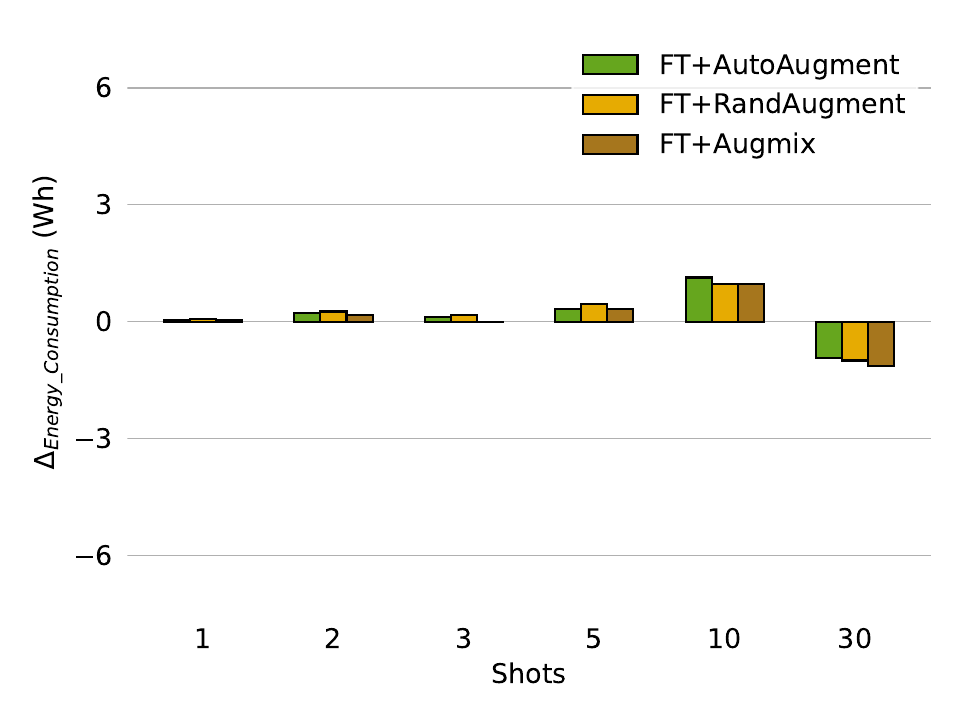}}
  \caption{Energy consumption difference between different DA strategies and vanilla finetuning for a varying number of shots.}
  \label{delta_energy_consumption}
\end{figure*}

Fig. \ref{delta_energy_consumption} illustrates the energy consumption difference between the automated DA selection techniques and the vanilla \texttt{FT} approach. Since all methods achieve the same $AP_{50}$ performance in all cases, this provides a valid comparison to determine the optimal strategy among the three. Following our previous observations, for the PPE dataset, automated DA selection methods result in increased energy consumption compared to \texttt{FT}, except for the 30-shot case, where the finetuning procedure converges faster. For the Fire dataset, there appears to be a trend where increasing the number of shots leads to a growing disparity between automated approaches and \texttt{FT}, with the only exception being the 5-shot case. In the CS dataset, the energy consumption overhead from using these automated approaches is relatively small. Overall, it is clear that these approaches do not exhibit consistent behavior across datasets, indicating that the data being used has a significant impact on their energy efficiency.

\subsection{Finetuning with and without DAs}

In the final assessment under the FSL scenario, we compare the performance of the best custom and automated DA approaches to the vanilla \texttt{FT} in terms of the $EF$ metric. Table \ref{ef_percent_change} shows the percent change between the $EF$ of \texttt{FT+DA(1)} (best custom DA technique in terms of $EF$) and \texttt{FT+AugMix} (best automated DA selection technique in terms of $EF$) compared to the $EF$ value of \texttt{FT}. Interestingly, for the PPE dataset, DA approaches lead to worse $EF$ performance, except \texttt{FT+AugMix} in the 30-shot scenario. In the Fire dataset, \texttt{FT+DA(1)} leads to a significant boost, especially when the number of shots is small. However, for a larger number of shots, that boost becomes small, and eventually, DA approaches perform worse than \texttt{FT} in the 30-shot scenario. Lastly, the benefits of utilizing \texttt{FT+AugMix} in the CS dataset are clearly illustrated, supported by the fact that the improvement upon the \texttt{FT} baseline generally increases as the number of shots increases. Consequently, there are no strong indications that DA approaches are always beneficial for FSL when considering both model performance and energy efficiency.

\begin{table}[]
\centering
\caption{Percent change of the Efficiency Factor (EF) when enhancing simple finetuning with DA strategies in the FSL scenario. Best results are indicated with \textcolor{blue}{blue(-)} when they are worse than the baseline and with \textcolor{red}{red(+)} when they are better.}
\label{ef_percent_change}
\resizebox{0.99\columnwidth}{!}{%
\begin{tabular}{llrrrrrr}
\toprule
\multicolumn{1}{l}{\multirow{2}{*}{\textbf{Dataset}}} & \multicolumn{1}{l}{\multirow{2}{*}{\textbf{Model}}} & \multicolumn{5}{c}{\textbf{Shots}}                                      \\ \cmidrule(l){3-8}
        & & \multicolumn{1}{c}{\textbf{1}} & \multicolumn{1}{c}{\textbf{2}} & \multicolumn{1}{c}{\textbf{3}} & \multicolumn{1}{c}{\textbf{5}} & \multicolumn{1}{c}{\textbf{10}} & \multicolumn{1}{c}{\textbf{30}} \\
\midrule
\multicolumn{1}{l}{\multirow{2}{*}{\textbf{PPE}}}
& FT+DA(1) & -17.32\%  & \textcolor{blue}{\textbf{-10.09\%}}  & \textcolor{blue}{\textbf{-8.33\%}}  & -9.72\% & -41.41\% & -49.87\% \\
& FT+AugMix\cite{hendrycks2019augmix} & \textcolor{blue}{\textbf{-8.22\%}} & -24.95\% & -23.06\% & \textcolor{blue}{\textbf{-6.39\%}} & \textcolor{blue}{\textbf{-3.70\%}} & \textcolor{red}{\textbf{+85.26\%}} \\
\midrule
\multicolumn{1}{l}{\multirow{2}{*}{\textbf{Fire}}} 
& FT+DA(1) & \textcolor{red}{\textbf{+55.35\%}} & \textcolor{red}{\textbf{+22.63\%}} & \textcolor{blue}{\textbf{-12.83\%}} & \textcolor{red}{\textbf{+50.33\%}} & \textcolor{red}{\textbf{+6.28\%}} & -22.53\% \\
& FT+AugMix\cite{hendrycks2019augmix} & +9.43\% & -27.89\% & -34.96\% & -1.31\% & -28.02\% & \textcolor{blue}{\textbf{-7.10\%}} \\
\midrule
\multicolumn{1}{l}{\multirow{2}{*}{\textbf{CS}}}
& FT+DA(1) & -5.40\% & -6.06\% & -6.91\% & -11.91\% & -10.23\% & -65.41\% \\
& FT+AugMix\cite{hendrycks2019augmix} & \textcolor{red}{\textbf{+3.49\%}} & \textcolor{red}{\textbf{+2.92\%}} & \textcolor{red}{\textbf{+9.62}} & \textcolor{red}{\textbf{+4.28\%}} & \textcolor{red}{\textbf{+44.27\%}} & \textcolor{red}{\textbf{+52.13\%}} \\
\bottomrule
\end{tabular}
}
\end{table}

\section{Low-Shot Learning Results} \label{LSL}
Unlike the FSL setting, where training tasks are constructed based on the number of bounding box annotations available for each class during training, a percentage of the total available training images is used for finetuning in the LSL setting. We define scenarios where $D\%$ of the available training images are used, including all bounding boxes found in these images, with $D \in \{5,20,25,50,75,100\}$. In the LSL scenario, we mainly focus on the proposed custom DA approaches and thus have included \texttt{FT+DA(1)}, which was particularly effective in the FSL scenario and \texttt{FT+DA(1+2)} to obtain a full view of utilizing the custom DAs altogether. Finally, we also include the results from finetuning a randomly initialized YOLOv8n model (results omitted in the FSL scenario since the model could not converge due to the very small number of training samples).

\begin{table}[]
\centering
\caption{Test set $AP_{50}$ for different training strategies and number of samples in the LSL scenario.}
\label{lsl_map}
\resizebox{0.9\columnwidth}{!}{%
\begin{tabular}{llrrrrrr}
\toprule
\multicolumn{1}{c}{\multirow{2}{*}{\textbf{Dataset}}} & \multicolumn{1}{c}{\multirow{2}{*}{\textbf{Model}}} & \multicolumn{5}{c}{\textbf{Data (\%)}}                                      \\ \cmidrule(l){3-8}
        & & $\mathbf{5\%}$ & $\mathbf{20\%}$ & $\mathbf{25\%}$ & $\mathbf{50\%}$ & $\mathbf{75\%}$ & $\mathbf{100\%}$ \\
\midrule
\multicolumn{1}{l}{\multirow{4}{*}{\textbf{PPE}}}
& Base & 10.19 & 34.61 & 36.45 & 50.23 & 48.28 & 59.18 \\
& FT & \textbf{43.20} & 58.80 & 57.80 & 69.60 & 73.30 & 73.40 \\
& FT+DA(1) & 34.21 & 41.05 & 44.63 & 50.16 & 47.83 & 58.60 \\
& FT+DA(1+2) & 36.80 & \textbf{92.40} & \textbf{80.10} & \textbf{95.10} & \textbf{84.70} & \textbf{96.40} \\
\midrule
\multicolumn{1}{l}{\multirow{4}{*}{\textbf{Fire}}} 
& Base & 15.13 & 25.40 & 27.91 & 39.76 & 37.52 & 37.59 \\
& FT & 25.30 & 41.20 & 35.60 & 39.90 & 46.00 & 53.50 \\
& FT+DA(1) & \textbf{43.84} & \textbf{57.98} & \textbf{59.69} & 56.93 & 57.45 & \textbf{91.79} \\
& FT+DA(1+2) & 40.70 & 51.00 & 48.50 & \textbf{58.60} & \textbf{71.00} & 87.50 \\
\midrule
\multicolumn{1}{l}{\multirow{4}{*}{\textbf{CS}}} 
& Base  & 30.12 & 53.65 & 56.16 & 67.70 & 69.85 & 74.54 \\
& FT & \textbf{63.70} & \textbf{76.30} & 73.80 & 75.40 & 75.00 & 81.50 \\
& FT+DA(1) & 41.49 & 49.92 & 54.96 & 50.79 & 60.83 & 61.44  \\
& FT+DA(1+2) & 61.80 & 73.40 & \textbf{76.00} & \textbf{83.60} & \textbf{83.40} & \textbf{86.20} \\
\bottomrule
\end{tabular}
}
\end{table}

\begin{figure*}[htp]
  \centering
  \subfigure[PPE Dataset]{\includegraphics[scale=0.33]{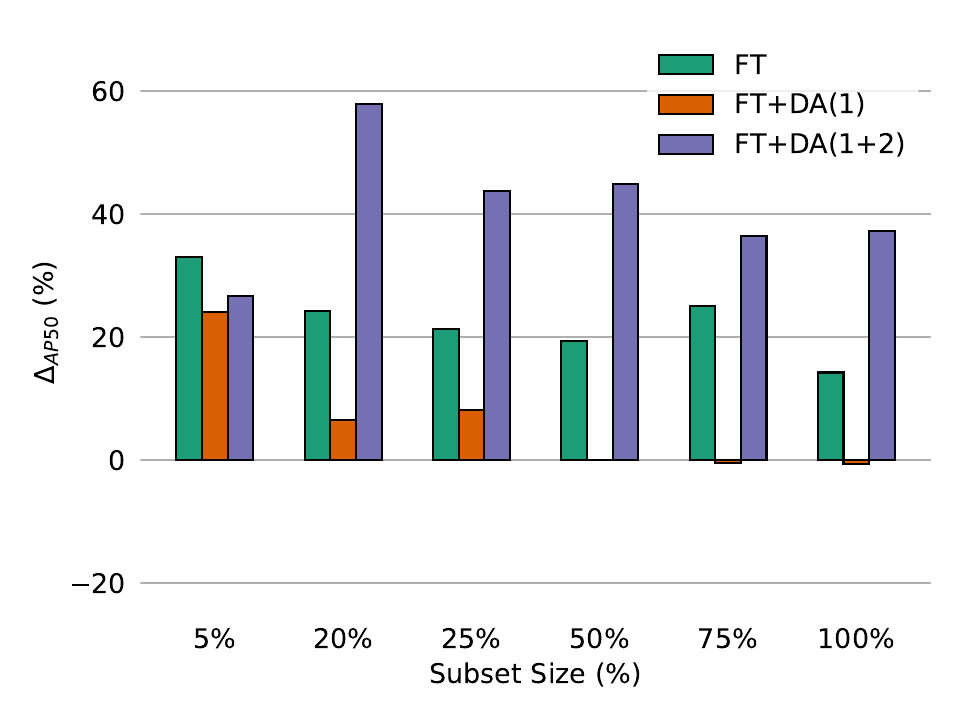}}
  \subfigure[Fire Dataset]{\includegraphics[scale=0.33]{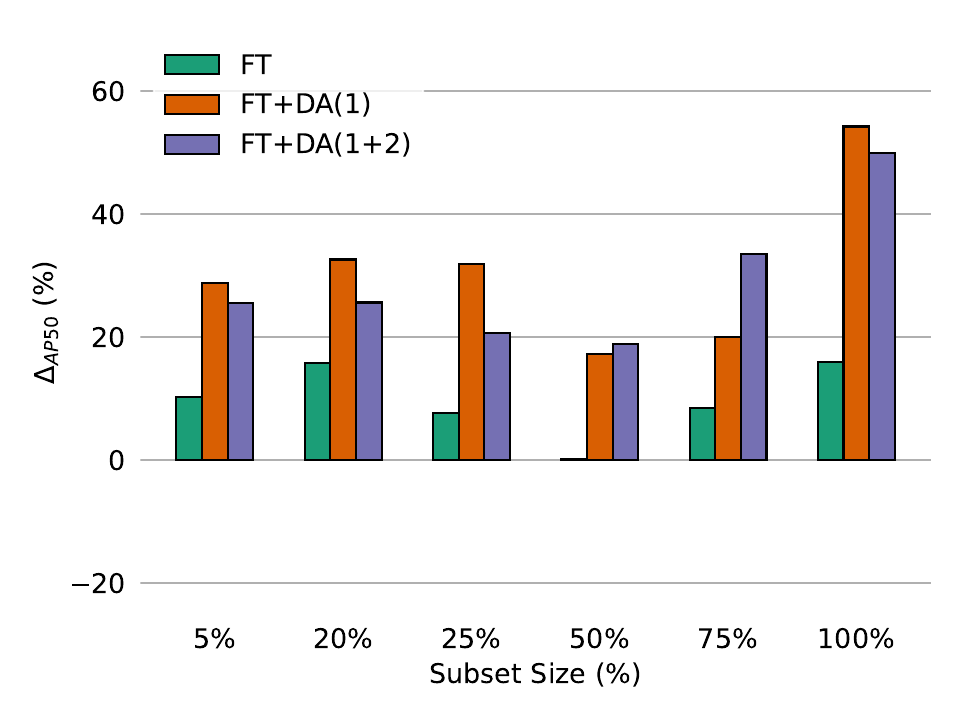}}
  \subfigure[CS Dataset]{\includegraphics[scale=0.33]{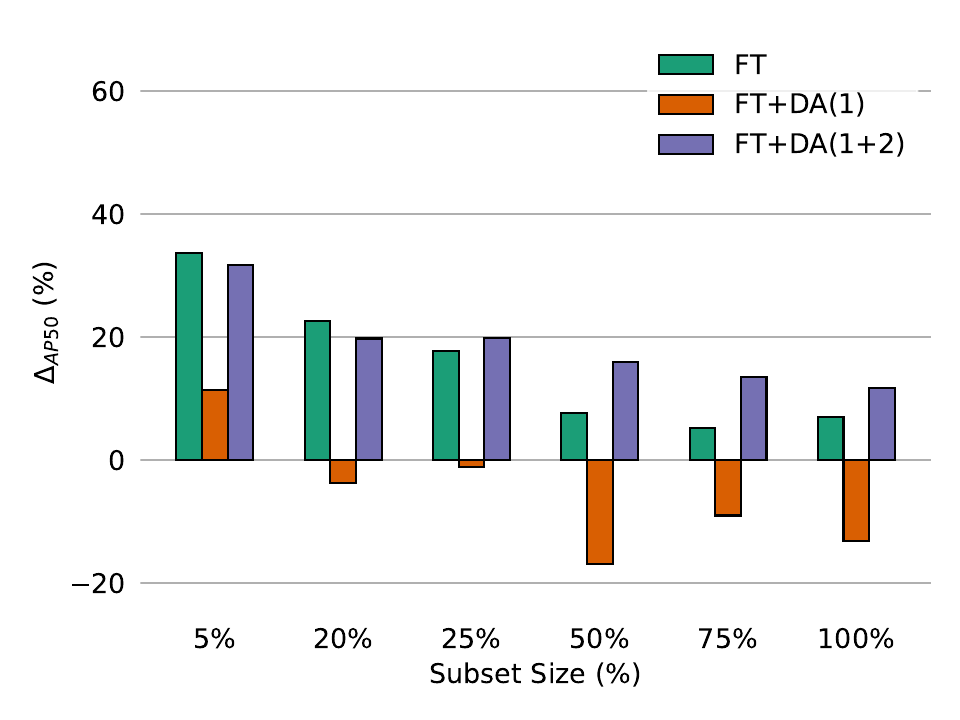}}
  \caption{$AP_{50}$ performance difference between finetuning-based approaches and \texttt{Base}.}
  \label{lsl_map_difference}
\end{figure*}

\subsection{Main Results}

Table \ref{lsl_map} includes the model performance of the examined methods under varying sizes of the training set in each dataset. In the PPE and CS datasets, vanilla \texttt{FT} leads to the best performance when the percentage of utilized data is small (5\% for PPE and 5\% and 20\% for CS), followed by \texttt{FT+DA(1+2)}. However, when the number of available training data increases, \texttt{FT+DA(1+2)} leads to optimal performance results in both cases. As for the Fire dataset, the application of custom DA approaches leads to improved results regardless of the training set size with \texttt{FT+DA(1+2)} achieving the best results for 5\%, 20\%, 25\% and 100\% of training data used and \texttt{FT+DA(1)} for 50\% and 75\%. Furthermore, it is clear that \texttt{Base} yields poor results in all cases, demonstrating the importance of combining pretraining and finetuning. This is also illustrated in Fig. \ref{lsl_map_difference} where using finetuning-based approaches can lead to significant improvements in $AP_{50}$ that exceed 50\% in some cases. However, in the CS dataset, using \texttt{FT+DA(1)} can lead to performance degradation when more than 5\% of the available training data is used, possibly due to the fact that \texttt{FT+DA(1)} introduces novel diverse training samples that cannot sufficiently be exploited by lightweight object detectors such as YOLOv8n \cite{tu2023femtodet}.

Table \ref{lsl_energy} contains the energy consumption during the finetuning process for each model in each dataset. It is important to note that training a model from scratch is highly inefficient compared to the finetuning approach in all cases, with energy requirements being three to four orders of magnitude higher. Overall, vanilla \texttt{FT} demonstrates the lowest energy consumption, especially as the number of available training data increases. When the percentage of training data is small, utilizing custom DA approaches can result in faster model convergence and thus lower energy consumption, as is the case for \texttt{FT+DA(1)} for 5\%, 20\%, and 25\% in the PPE dataset and \texttt{FT+DA(1)} for 5\% in the Fire dataset. Additionally, similar to the findings in the FSL case, energy consumption generally increases as the number of available training samples increases.

\begin{table}[]
\centering
\caption{Total energy consumption in Wh during training for different training strategies and number of samples in the LSL scenario.}
\label{lsl_energy}
\resizebox{\columnwidth}{!}{%
\begin{tabular}{llllllll}
\toprule
\multicolumn{1}{l}{\multirow{2}{*}{\textbf{Dataset}}} & \multicolumn{1}{l}{\multirow{2}{*}{\textbf{Model}}} & \multicolumn{5}{c}{\textbf{Data (\%)}}                                      \\ \cmidrule(l){3-8}
        & & \multicolumn{1}{c}{$\mathbf{5\%}$} & \multicolumn{1}{c}{$\mathbf{20\%}$} & \multicolumn{1}{c}{$\mathbf{25\%}$} & \multicolumn{1}{c}{$\mathbf{50\%}$} & \multicolumn{1}{c}{$\mathbf{75\%}$} & \multicolumn{1}{c}{$\mathbf{100\%}$} \\
\midrule
\multicolumn{1}{l}{\multirow{4}{*}{\textbf{PPE}}}
& Base & {$4.40\times10^4$} & {$5.13\times10^4$} & {$5.46\times10^4$} & {$5.06\times10^4$} & {$8.42\times10^4$} & {$6.79\times10^4$} \\
& FT & {$1.99\times10^1$} & {$2.72\times10^1$} & {$2.37\times10^1$} & {$2.52\times10^1$} & {$\mathbf{3.61\times10^1}$} & {$\mathbf{4.44\times10^1}$} \\
& FT+DA(1) & {$\mathbf{9.44\times10^0}$} & {$\mathbf{1.46\times10^1}$} & {$\mathbf{1.40\times10^1}$} & {$4.42\times10^1$} & {$8.19\times10^1$} & {$1.24\times10^2$} \\
& FT+DA(1+2) & {$1.06\times10^1$} & {$3.65\times10^1$} & {$2.05\times10^1$} & {$\mathbf{2.44\times10^1}$} & {$3.85\times10^1$} & {$6.16\times10^1$} \\
\midrule
\multicolumn{1}{l}{\multirow{4}{*}{\textbf{Fire}}} 
& Base & {$9.97\times10^4$} & {$1.83\times10^5$} & {$1.81\times10^5$} & {$3.47\times10^5$} & {$2.80\times10^5$} & {$3.41\times10^5$} \\
& FT & {$3.49\times10^1$} & {$\mathbf{3.55\times10^1}$} & {$\mathbf{3.19\times10^1}$} & {$\mathbf{5.80\times10^1}$} & {$\mathbf{1.32\times10^2}$} & {$\mathbf{1.62\times10^2}$} \\
& FT+DA(1) & {$\mathbf{3.05\times10^1}$} & {$1.78\times10^2$} & {$1.09\times10^2$} & {$2.13\times10^2$} & {$3.24\times10^2$} & {$1.14\times10^3$} \\
& FT+DA(1+2) & {$1.13\times10^2$} & {$2.18\times10^2$} & {$1.07\times10^2$} & {$3.49\times10^2$} & {$6.84\times10^2$} & {$6.38\times10^2$} \\
\midrule
\multicolumn{1}{l}{\multirow{4}{*}{\textbf{CS}}} 
& Base & {$4.63\times10^4$} & {$1.08\times10^5$} & {$1.17\times10^5$} & {$1.52\times10^5$} & {$2.66\times10^5$} & {$3.53\times10^5$} \\
& FT & {$\mathbf{8.04\times10^0}$} & {$\mathbf{1.37\times10^1}$} & {$\mathbf{1.52\times10^1}$} & {$\mathbf{2.52\times10^1}$} & {$\mathbf{3.61\times10^1}$} & {$\mathbf{4.44\times10^1}$} \\
& FT+DA(1) & {$1.45\times10^1$} & {$3.71\times10^1$} & {$4.67\times10^1$} & {$8.30\times10^1$} & {$1.21\times10^2$} & {$1.61\times10^2$} \\
& FT+DA(1+2) & {$1.33\times10^1$} & {$3.57\times10^1$} & {$4.61\times10^1$} & {$8.45\times10^1$} & {$1.24\times10^2$} & {$1.71\times10^2$} \\
\bottomrule
\end{tabular}
}
\end{table}

Finally, Table \ref{lsl_ef} displays the $EF$ metric values of the examined approaches at varying percentages of training data in each dataset. While using DAs in the LSL setting can lead to better performance, when considering each method's performance and energy efficiency together, \texttt{FT} yields the best results. This is especially evident in the Fire and CS datasets, where the method with the lowest energy consumption achieves the highest $EF$ value, similar to the FSL setting. However, in the PPE dataset, the performance improvements from using \texttt{FT+DA(1)} and \texttt{FT+DA(1+2)} are so significant that they outweigh the energy consumption overhead introduced by these DAs. Consequently, their $EF$ values are favorable compared to \texttt{FT}, except when using 100\% training data (where \texttt{FT+DA(1+2)} is only slightly inferior to \texttt{FT}).

\begin{table}[]
\centering
\caption{Efficiency factor (EF) metric values for different training strategies and number of samples in the LSL scenario.}
\label{lsl_ef}
\resizebox{0.9\columnwidth}{!}{%
\begin{tabular}{llrrrrrr}
\toprule
\multicolumn{1}{l}{\multirow{2}{*}{\textbf{Dataset}}} & \multicolumn{1}{l}{\multirow{2}{*}{\textbf{Model}}} & \multicolumn{5}{c}{\textbf{Data (\%)}}                                      \\ \cmidrule(l){3-8}
        & & {$\mathbf{5\%}$} & {$\mathbf{20\%}$} & {$\mathbf{25\%}$} & {$\mathbf{50\%}$} & {$\mathbf{75\%}$} & {$\mathbf{100\%}$} \\
\midrule
\multicolumn{1}{l}{\multirow{4}{*}{\textbf{PPE}}}
& {Base ($\times10^{-4}$)} & 2.316 & 6.747 & 6.676 & 9.927 & 5.734 & 8.716 \\
& FT & 2.067 & 2.085 & 2.340 & 2.657 & 1.976 & \textbf{1.617} \\
& FT+DA(1) & \textbf{3.277} & \textbf{2.631} & 2.975 & 1.110 & 0.577 & 0.469 \\
& FT+DA(1+2) & 3.172 & 2.464 & \textbf{3.726} & \textbf{3.744} & \textbf{2.144} & 1.540 \\
\midrule
\multicolumn{1}{l}{\multirow{4}{*}{\textbf{Fire}}} 
& {Base ($\times10^{-4}$)} & 1.518 & 1.388 & 1.542 & 1.146 & 1.340 & 1.102 \\
& FT & 0.705 & \textbf{1.129} & \textbf{1.082} & \textbf{0.676} & \textbf{0.346} & \textbf{0.328} \\
& FT+DA(1) & \textbf{1.392} & 0.324 & 0.543 & 0.266 & 0.177 & 0.080 \\
& FT+DA(1+2) & 0.357 & 0.233 & 0.449 & 0.167 & 0.104 & 0.137 \\
\midrule
\multicolumn{1}{l}{\multirow{4}{*}{\textbf{CS}}} 
& {Base ($\times10^{-4}$)} & 6.505 & 4.968 & 4.800 & 4.454 & 2.626 & 2.112 \\
& FT & \textbf{7.047} & \textbf{5.191} & \textbf{4.556} & \textbf{2.878} & \textbf{2.022} & \textbf{1.796} \\
& FT+DA(1) & 2.677 & 1.310 & 1.152 & 0.605 & 0.499 & 0.379 \\
& FT+DA(1+2) & 4.322 & 2.000 & 1.614 & 0.978 & 0.667 & 0.501 \\
\bottomrule
\end{tabular}
}
\end{table}

\subsection{Performance vs Efficiency Trade-offs in the LSL Setting}

While the results mentioned above suggest that using DA techniques can enhance model performance at the cost of increased energy consumption, it remains uncertain whether this trade-off is beneficial when taking both performance and energy efficiency into account. To address this question, Table \ref{ef_percent_change_lsl} includes the percent change of the $EF$ metric values between the best custom DA approaches and vanilla \texttt{FT} for the examined LSL scenarios. Interestingly, utilizing DA methods is optimal only in the PPE dataset, mainly because they significantly improve $AP_{50}$. However, even in this scenario, as the number of available training samples increases, the advantages of using custom DAs diminish and vanish entirely in the case of 100\% available data. For the Fire and CS datasets, it is apparent that in most cases, using DAs leads to significantly worse results compared to \texttt{FT}. Overall, this raises the question of whether DAs in the LSL setting are actually advantageous when considering both performance and energy efficiency.

\begin{table}[]
\centering
\caption{Percent change of the Efficiency Factor (EF) when enhancing simple finetuning with DA strategies in the LSL scenario. Best results are indicated with \textcolor{blue}{blue(-)} when they are worse than the baseline and with \textcolor{red}{red(+)} when they are better.}
\label{ef_percent_change_lsl}
\resizebox{0.99\columnwidth}{!}{%
\begin{tabular}{llrrrrrr}
\toprule
\multicolumn{1}{l}{\multirow{2}{*}{\textbf{Dataset}}} & \multicolumn{1}{l}{\multirow{2}{*}{\textbf{Model}}} & \multicolumn{5}{c}{\textbf{Data (\%)}}                                      \\ \cmidrule(l){3-8}
        & & \multicolumn{1}{c}{\textbf{5\%}} & \multicolumn{1}{c}{\textbf{20\%}} & \multicolumn{1}{c}{\textbf{25\%}} & \multicolumn{1}{c}{\textbf{50\%}} & \multicolumn{1}{c}{\textbf{75\%}} & \multicolumn{1}{c}{\textbf{100\%}} \\
\midrule
\multicolumn{1}{l}{\multirow{2}{*}{\textbf{PPE}}}
& FT+DA(1) & \textcolor{red}{\textbf{+58.54\%}} & \textcolor{red}{\textbf{+26.19\%}} & +27.14\% & -58.22\% & -70.80\% & -71.00\% \\
& FT+DA(1+2) & +53.46\% & +18.18\% & \textcolor{red}{\textbf{+59.23\%}} & \textcolor{red}{\textbf{+40.91\%}} & \textcolor{red}{\textbf{+8.50\%}} & \textcolor{blue}{\textbf{-4.76\%}} \\
\midrule
\multicolumn{1}{l}{\multirow{2}{*}{\textbf{Fire}}} 
& FT+DA(1) & \textcolor{red}{\textbf{+97.45\%}} & \textcolor{blue}{\textbf{-71.30\%}} & \textcolor{blue}{\textbf{-49.82\%}} & \textcolor{blue}{\textbf{-60.65\%}} & \textcolor{blue}{\textbf{-48.84\%}} & -75.61\% \\
& FT+DA(1+2) & -49.36\% & -79.36\% & -58.50\% & -75.30\% & -69.94\% & \textcolor{blue}{\textbf{-58.23\%}} \\
\midrule
\multicolumn{1}{l}{\multirow{2}{*}{\textbf{CS}}}
& FT+DA(1) & -62.01\% & -74.76\% & -74.71\% & -78.98\% & -75.32\% & -78.90\% \\
& FT+DA(1+2) & \textcolor{blue}{\textbf{-38.67\%}} & \textcolor{blue}{\textbf{-61.47\%}} & \textcolor{blue}{\textbf{-64.57\%}} & \textcolor{blue}{\textbf{-66.01\%}} & \textcolor{blue}{\textbf{-67.01\%}} & \textcolor{blue}{\textbf{-72.10\%}} \\
\bottomrule
\end{tabular}
}
\end{table}

\section{Conclusion}
\label{sec:Conclusion}
% V0.1
% While low/few-shot object detection has been extensively studied in recent years, the performance impact of different data augmentation strategies on finetuning-based models, as well as their energy efficiency in this context, have not been fully explored. This paper aims to address these gaps by conducting an empirical study that evaluates the performance and energy consumption of various data augmentation strategies in data-scarce settings. Using a lightweight object detector finetuned on downstream tasks with limited data as a baseline, we assessed its performance and efficiency when enhanced with custom augmentations and automated data augmentation selection algorithms. Our evaluations in three challenging industrial object detection tasks demonstrate that while data augmentations can improve performance in these data regimes, the additional energy consumption often outweighs the performance gains. Consequently, the Efficiency Factor metric is employed to assess both model performance and energy efficiency, concluding that the effectiveness of data augmentations highly depends on the dataset and may not always lead to improved results. In the future, it would be interesting to explore how these insights can be utilized in designing augmentation strategies that consider both model performance and energy efficiency in these settings.

% V0.2
While low/few-shot object detection has been extensively studied in recent years, the performance impact of different data augmentation strategies on finetuning-based models and their energy efficiency have yet to be fully explored. This paper aims to address these gaps by conducting an empirical study that evaluates data augmentation strategies in data-scarce settings and their effect on a lightweight object detector's performance and energy consumption. Results on three challenging industrial object detection tasks with limited data show that while data augmentations can improve performance, the additional energy consumption often outweighs the performance gains. Finally, a novel Efficiency Factor metric is employed to assess both model performance and energy efficiency, concluding that the effectiveness of data augmentations highly depends on the dataset and may not always lead to improved results. In the future, it would be interesting to explore how these insights can be utilized in designing augmentation strategies that enhance both model performance and energy efficiency in these settings as well as other application domains with similar requirements, e.g., healthcare and the energy sector.

\bibliographystyle{IEEEtran}
\bibliography{bibliography.bib}

\end{document}